# Speech Recognition: Increasing Efficiency of Support Vector Machines

### Aamir Khan
COMSATS Insititute of Information Technology,

WahCantt.
Pujnab, Pakistan

### Muhammad Farhan
University of Science and Technology Bannu
KPK, Pakistan

### Asar Ali
City University of Science and Information Technology,
Peshawar, KPK, Pakistan

## ABSTRACT
With the advancement of communication and security technologies, it has become crucial to have robustness of embedded biometric systems. This paper presents the realization of such technologies which demands reliable and error-free biometric identity verification systems. High dimensional patterns are not permitted due to eigen-decomposition in high dimensional feature space and degeneration of scattering matrices in small size sample. Generalization, dimensionality reduction and maximizing the margins are controlled by minimizing weight vectors. Results show good pattern by multimodal biometric system proposed in this paper. This paper is aimed at investigating a biometric identity system using Support Vector Machines(SVMs) and Lindear Discriminant Analysis(LDA) with MFCCs and implementing such system in real-time using SignalWAVE.

## 1. INTRODUCTION
The Performance of a speech recognition system is affected considerably by the choice of features for most of the applications. Thus, it is important to find an appropriate representation of such continuous speech. Raw data obtained from speech recording can't be used to directly train the recognizer as for the same phonemes do not necessarily have same sample values. Features which are based on signal spectra are of better choice. Our representation should be compact and features must add information to the recognition process to satisfy the property of features being independent. Here we proposed a system which increases the efficiency of support vector machines by apply this technique after LDA implementation.

- ➢ MFCC
- ➢ LDA
- ➢ SVM

**MFCC** A block diagram of the structure of an MFCC processor is shown in the figure 1.1 below. The speech input is typically recorded. The sampling frequency 10000 Hz is chosen to minimize the effects of aliasing in the analog-to-digital conversion. In addition, rather than the speech waveforms themselves, Mel-Frequency Cepstrum Coefficient processor is known to be less susceptible to already mentioned variations.

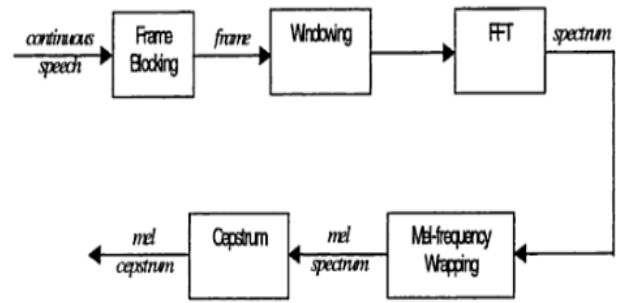

**Fig** 1.1 Block diagram of the MFCC processor

Frame Blocking is the process of segmenting the speech samples obtained from analog to digital conversion (ADC) into a small frame with the length within the range of 20 to 40 msec. Hamming windowing is employed to window each individual frame so as to minimize the signal discontinuities at the beginning and end of each frame. The concept here is to minimize the spectral distortion by using the window to taper the signal to zero at the beginning and end of each frame. Fast Fourier Transform converts each frame of N samples from the time domain into the frequency domain.

For each tone with an actual frequency, f, measured in Hz, a subjective pitch is measured on a scale called the 'mel' scale. The mel-frequency scale is linear frequency spacing below 1000 Hz and a logarithmic spacing above 1000 Hz.

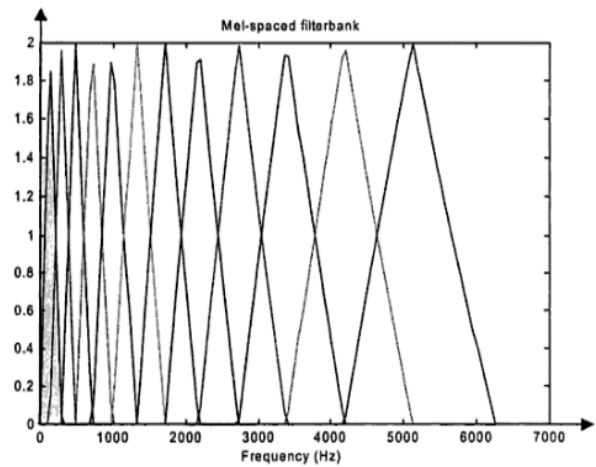

**Fig** 1.2 An example of mel-spaced filterbank

The subjective spectrum is to use a filter bank, spaced uniformly on the mel-scale (see Figure 1.2).

The log mel spectrum is converted back to time. The result is called the mel frequency cepstrum coefficients (MFCC).

**LDA** Linear Discriminent Analysis is a statistical method which reduces the dimension of the features while maximizing the information preserved in the reduced feature space. Use of lda after mfcc drastically reduces the dimension of features as LDA finds optimal transformation matrix which preserve most of the information and the same can be used to discriminate between the different classes. Further it is necessary to use a label of the recorded data as to associate each speech segment with a label.

**SVM** Support Vector Machine i.e. SVMs were introduced by Boser, Guyon and Vapnik in COLT-92 [4-6], works on principle of based on some previous training through inputs, using supervised learning techniques to classify data [7]. Using machine learning theory to boost higher accuracy and avoid the data to be over-fit automatically, SVM is classifier. SVM can be thought of systems using hyper planes i.e. hypothesis space of linear functions in feature space of high dimensions. Hyper planes are trained with a learning algorithm to optimize and use statistical learning based learning bias. SVM has been tested for hand writing, face recognition in general pattern classification and regression based applications. Even though having complex hierarchy and design SVM provide good results than neural networks. SVM performing prediction tasks on data instances in the testing instances. SVM method of classification imitates supervised learning involving identification referred as feature extraction and produce favorable outputs. The biggest advantage of SVM is easy to train and scale complex high dimensional data as compared to neural networks at the expense kernel function to guide the SVM [2, 12]. SVM can be applied to regression by introducing an alternative loss function [11, 12].

## 2. METHODOLOGY

This paper proposes a three stage design. First stage is about getting the recorded data and then passing it through the MFCC processor. Frame Blocking is blocking speech signals into frames of N samples, with adjacent frames being separated by M (M < N). The first frame consists of the first N samples. The second frame begins M samples after the first frame, and overlaps it by N - M samples and so on. This process continues until all the speech is accounted for within one or more frames. Typical values for N and M are N = 256 (which is equivalent to ~ 30 msec windowing and facilitate the fast radix-2 FFT) and M = 100.

The result of windowing is the signal

$$y_l(n) = x_l(n)h(n), \quad 0 \le n \le N-1$$

Typically the Hamming window is used, which has the form:

$$h(n) = 0.54 - 0.46\cos\left(\frac{2\pi n}{N-1}\right), \quad 0 \le n \le N-1$$

Transforming into: h(n)0.54 – 0.46*cos(2*pi*(0:windowSize-1)/windowSize) as used in the MATLAB code. Fast Fourier Transform converts each frame of $N$ samples from the time domain into the frequency domain.

$$X_k = \sum_{n=0}^{N-1} x_n e^{-j2\pi kn/N}, \qquad k = 0,1,2,...,N-1$$

The result after this step is often referred to as spectrum. That filter bank has a triangular bandpass frequency response, and the spacing as well as the bandwidth is determined by a constant mel frequency interval. The number of mel spectrum coefficients, $K$, is typically chosen as 20.

Because the mel spectrum coefficients (and so their logarithm) are real numbers, we can convert them to the time domain using the Discrete Cosine Transform (DCT). Therefore if we denote those mel power spectrum coefficients that are the result of the last step are $\tilde{S}_0, k = 0,2,...,K-1$, we can calculate the MFCC's, $\tilde{c}_n$, as

$$\tilde{c}_n = \sum_{k=1}^{K}(\log \tilde{S}_k)\cos\left[n\left(k-\frac{1}{2}\right)\frac{\pi}{K}\right], \qquad n = 0,1,...,K-1$$

Where n=1,2,....K

Note that we exclude the first component, $\tilde{c}_0$, from the DCT since it represents the mean value of the input signal, which carried little speaker specific information.

In order to reduce the large dimension of the features, most discriminating speech features are further acquired by using the LDA by calculating the Class Matrix, Between Class Matrix, Generalized Eigenvalue and Eigenvector, sorting the order of eigenvector and projecting training records onto the fisher basis vectors.

$$S_i = \sum_{x \in X_i}(x - m_i)(x - m_i)^{\mathrm{T}}$$
$$S_W = \sum_{i=1}^{C} S_i$$

Where C is the number of classes

$$S_B = \sum_{i=1}^{C} n_i (x - m_i)(x - m_i)^{\mathrm{T}}$$

Where $n_i$ is the number of feature data in the $ith$ class, $m$ is the total mean of all the training recorded data.

For linear SVM, separating data with a hyper plane and extending to non-linear boundaries by using kernel trick [18, 12] and correct classification of all data is mathematically

i)      $wx_i + b \ge 1$,   if $y_i = +1$

ii)     $wx_i + b \le 1$,   if $y_i = -1$

iii)    $y_i(w_i + b) \ge 1$, for all i

Where w is a weight vector. Depending upon the quality of the training data to be good and every test vector in the range of radius r from the training vector, the hyper plane selected is considered at the farthest possible region from the data.Linear constraints are required to optimize the quadratic function

$$f(\mathbf{x}) = \Sigma \alpha_i y_i x_i * \mathbf{x} + \boldsymbol{b}$$

For data with linear attribute, a separating hyper plane suffice to divide the data, however, in most cases the data is non linear and inseparable. Kernels use the non-linearly mapped data into a space of high dimensions and distinguish the linear data [7]. For example in [12]

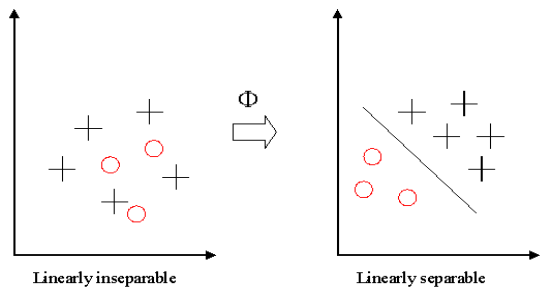

**Fig** 2.1

Kernel defines the mapping as

$$k(x, y) = \emptyset(x).\emptyset(y)$$

After transforming data into feature space, similarity measure can be defined on basis of dot product. A good choice of feature space makes the pattern recognition easier [7].

$$\langle x1.x2 \rangle \leftarrow K(x1, x2) = \langle \emptyset(x1).\emptyset(x2) \rangle$$

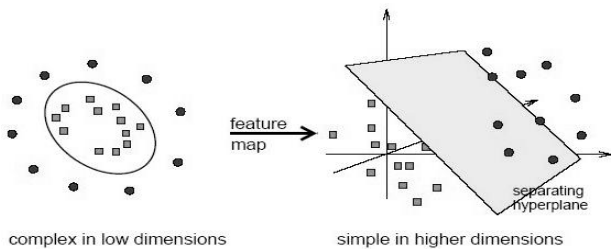

**Fig** 3.2 Feature Space Representation

MMFC is transformed into feature-vectors. Machine learning as a part of Artificial Intelligence helps machines in decision making[7]. Boser, Guyon and Vapnik in COLT-92 [4-6] introduced SVM in 1992 as linear classifiers. If pixel maps are used SVM give accurate results over complicated neural networks [2]. Neural networks use traditional Empirical Risk Minimization (ERM) but SVM use Structural Risk

Minimization (SRM) principle and show good results [8, 9]. SRM minimize the upper bound on error risk and ERM minimize the error risk of training data. SVM are used as classifiers but recently have been used for regression purposes [10].

The kernel trick allows SVM to extrapolate non-linear boundaries. Few steps need to be followed for kernal trick [7]:

Only the inner product of datasets are used for the expression of algorithm and referred as the dual problem.

- Actual data is accepted through a non-linear mapping algorithm to form a new set so data with respect to the newly formed dimensions by manipulating the data vectors through a pair wise product of some of the original data.
- For non-linearly mapped dot product can be represented without inner product on large vectors. This function is the kernel function.

Kernel functions enable operations into input space instead of complex high dimensional feature space. Based upon reproduction of Kernel Hilbert Spaces [12] inherently put the attributes of inputs into feature space. SVM generalize and approximate training data better on given data sets.SVM make it possible to determine parameters which control complexity of the system by performing Cross-Validation on available datasets. The Gaussian and polynomial kernels are the standard choice. The Mercer kernels provide an ideal feature mapping [11,13]. Similar techniques are used for the classification of text and image patterns.

After properly generating the Simulink block sets and editing the program to identify the various clients, the next task was that of implementing the design onto the SignalWAVE development board in order to test the DSP algorithms and perform hardware/software parallel design in real-time. The SignalWAVE development board provides faster implementation through Simulink-based DSP and FPGA code generation. These board is a complete DSP-FPGA-based, real-time development tool for next-generation, wireless audio-video telecommunication applications that integrates audio and video I/Os, as well as high-speed converters. The blocks were to be downloaded but a memory problem was encountered. The LDA and SVMs features were supposedly too large and the SignalWave FPGA board memory was not enough to process the speech signal. The data was taken directly from workspace and then it was finally downloaded.

## 3. RESULTS

The result for the sound file chosen from the testing sound and it was the 5th sound file as there are two test sound files for each word. Hence it is shown that SVM efficiency is reduced when the dimensionality of the training data is increased. It is shown that if LDA and SVM are both used together produce very good results as shown in Fig.1. And SVM to audio data before LDA and LDA projection train are shown in Fig. 2 and Fig. 3 respectively.

```
>> svmtrain(classlabel, ldaprojtrainingimg','-g 2 -c 10 -v 10')
Cross Validation Accuracy = 100%

ans =

     100

fx >> |
```

**Fig. 3.1:** SVM after LDA Cross-validation

```
>> svmtrain(classlabel, audiodata','-g 2 -c 10 -v 10')
Cross Validation Accuracy = 60%

ans =

      60

fx >> |
```

**Fig. 3.2:** SVM to Audio data before LDA

```
>> model= svmtrain(classlabel, ldaprojtrainingimg','-g 2 -c 10')

model =

    Parameters: [5x1 double]
      nr_class: 5
       totalSV: 5
           rho: [10x1 double]
         Label: [5x1 double]
         ProbA: []
         ProbB: []
           nSV: [5x1 double]
       sv_coef: [5x4 double]
           SVs: [5x5 double]

>> predict= svmpredict(classlabel2,ldaprojtestimage,model)
Accuracy = 40% (2/5) (classification)

predict =

     1
     1
     5
     5
     5

>> |
```

**Fig. 3.3:** LDA Projection train

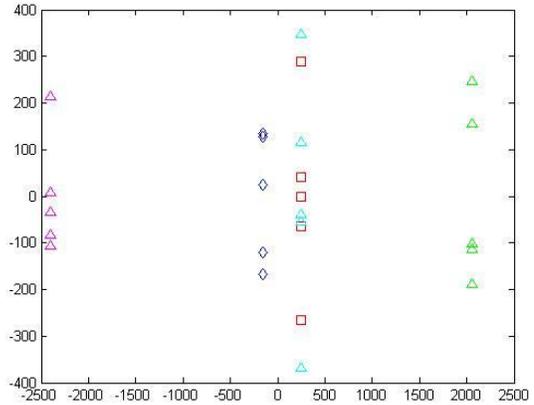

**Fig** 3.4    LDA plot.

After making each block in Simulink, constant referral was made to the workspace in order to match the output from simulink blocks to the output from the code. The final result which was obtained at the display block is shown below.

For K=2

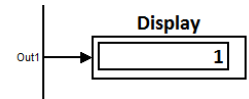

**Fig 3.5** Result for k=2 (test database)

For K=5:

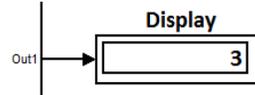

**Fig 3.6** Result for k=5 (test database)

For K=10:

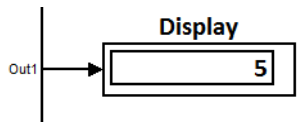

**Fig 3.7** Result for k=10 (test database)

## 4. CONCLUSION AND FUTURE WORK

Results show how SVM alone is not sufficient for classification and poor results are observed. However, SVM when used with statistical tool of LDA which is used to reduce the dimension of the features data received from MFCC processor increases the efficiency and proves to be 100% accurate and the prediction rate for more than 40%. It is shown here SVM efficiency is reduced when the dimensionality of the training data is increased. The paper also delineates a feasible solution for implementing the proposed system on FPGA for significant speed increase. The efficiency highly increases with the use of SVM over LDA. Speech recognition system proposed in this paper based on real tests give accurate results and significantly decreases the complexity. It is observed during the whole study that if SVM and LDA are both used jointly it produce accurate results. Furthermore, it is recommended to experiment HLDA with support vector machine (SVM) for speech recognition.

# 5. REFERENCES

[1] Belhumeur P., Hespanha J., and Kriegeman D., "EigenfacesvsFisherfaces", 1997.

[2] Recognition Using Class Specific Linear Projection, IEEE Trans. PAMI, 19(7): pp. 711-729.

[3] NelloCristianini and John Shawe-Taylor,"An Introduction to Support Vector Machines and other Kernel-based Learning Methods", Cambridge University Press, 2000.

[4] Burges C., "A Tutorial on Support Vector Machines for Pattern Recognition", In Data Mining and Knowledge Discovery, Vol. II, Kluwer Academic Publishers, Boston, 1998.

[5] V. Vapnik, S. Golowich, and A. Smola,"Support Vector Method for Function Approximation, Regression Estimation and Signal Processing", In M. Mozer, M. Jordan, and T. Petsche, (edited), Advances in Neural Information Processing Systems 9, pp.281–287,Cambridge, MA, MIT Press 1997.

[6] C.Cortes and V. Vapnik "Support Vector Networks. Machine Learning", 20: pp.273 – 297, 1995.

[7] N. Heckman. "The Theory and Application of Penalized Least Squares Methods for Reproducing Kernel Hilbert-spaces Made Easy",1997.

[8] David M Skapura, "Building Neural Networks", ACM press, 1996.

[9] M. Farhan, "Investigation of Support Vector Machine as Classifier",MS Thesis Nottingham University Malaysia Campus, 2010.

[10] Tom Mitchell, "Machine Learning", McGraw-Hill Computer Science Series, 1997.

[11] David M Skapura,, "Building Neural Networks", ACM Press, 1996.

[12] M. Aizerman, E. M. Braverman, and L.I.Rozono´er "Theoretical Foundations of Potential Function Method in Pattern Recognition Learning", Automation and Remote Control, 25: pp. 821–837, 1964.

[13] N. Aronszajn, "Theory of Reproducing Kernels", Transaction American Mathematical Society, 686: pp. 337-404, 1950.

[14] Anil K. Jain, Jianchang Mao, K. M. Mohiuddin, *Artificial Neural Networks: A Tutorial*, Computer, v.29 n.3, p.31-44, March 1996

[15] Simon Haykin, *Neural Networks: A comprehensive foundation*, 2nd Edition, Prentice Hall, 1998

[16] Alexander J. Faaborg, *Using Neural Networks to Create an Adaptive Character Recognition System*, March 2002

[17] E. W. Brown, *Character Recognition by Feature Point Extraction*, unpublished paper authored at Northeastern University, 1992, available at: http://www.ccs.neu.edu/home/fe neric/charrecnn.html

[18] S. Furui, "An overview of speaker recognition technology", *ESCA Workshop on Automatic Speaker Recognition,Identification and Verification*, pp. 1-9, 1994.

[19] F.K. Song, A.E. Rosenberg and B.H. Juang, "A vector quantisation approach to speaker recognition", *AT&T Technical Journal*, Vol. 66-2, pp. 14-26, March 1987.

[20] S. Balakrishnama, A. Ganapathiraju, Linear Discriminant Analyais - A Brief Tutorial, Institute for Signal and Information Processing, Department of Electrical andComputer Engineering, Mississippi State University, page 2-3.

[21] Bishop, C. (1995). *Neural Networks for Pattern Recognition*. Oxford: University Press. Extremely well-written, up-to-date.Requires a good mathematical background, but rewards careful reading, putting neural networks firmly into a statistical context.